\documentclass{article}

\usepackage[dblblindworkshop,final]{neurips_2026}
\usepackage{graphicx}
\usepackage{amsmath}
\usepackage[utf8]{inputenc}
\usepackage[T1]{fontenc}
\usepackage{hyperref}
\usepackage{url}
\usepackage{booktabs}
\usepackage{amsfonts}
\usepackage{nicefrac}
\usepackage{microtype}
\usepackage{xcolor}
\usepackage{soul}
\usepackage{placeins}

\title{COVERAGE-TRACE: Source-Side Counterfactual Tracing for Detecting Omissions in LLM Summaries}

\workshoptitle{Long-Context Foundation Models (LCFM)}

\author{%
  {\normalfont\bfseries\normalsize
  Ali Karkehabadi$^{1,3}$ \hspace{0.6em}
  Jamshid Hassanpour$^{2}$ \hspace{0.6em}
  Rafael E. Banchs$^{3,4}$}\\[5pt]
  {\normalfont\small
  $^{1}$University of California, Davis \quad
  $^{2}$Georgia Institute of Technology}\\[2pt]
  {\normalfont\small
  $^{3}$Intapp \quad
  $^{4}$San Jose State University}\\[5pt]
  {\normalfont\small
  $^{1}$\text{akarkehabadi@ucdavis.edu} \quad
  $^{2}$\text{jamshid@gatech.edu}}\\[2pt]
  {\normalfont\small
  $^{3}$\text{\{ali.karkehabadi,rafael.banchs\}@intapp.com}}\\[2pt]
  {\normalfont\small
  $^{4}$\text{rafael.banchs@sjsu.edu}}%
}

\begin{document}

\maketitle

\begin{abstract}
Omissions are difficult to verify because missing information leaves no output
span to inspect: a summary may be factually supported while still excluding a
decision, qualification, quantity, or condition needed to preserve the source's
meaning. We introduce \textbf{COVERAGE-TRACE}, a source-side counterfactual
framework that evaluates omission as dependence on a designated atomic source
fact. It replaces only the target fact with type-matched alternatives, holds the
candidate summary fixed, and measures the resulting change in its
teacher-forced, length-normalized likelihood. A summary covering the target
should respond more strongly than one omitting it, while aggregation across
multiple counterfactuals reduces sensitivity to replacement choice.
We evaluate COVERAGE-TRACE on 80 matched covered--omitted QMSum pairs spanning
several fact types and four open-weight models. Multi-CF achieves
0.764--0.825 AUROC, compared with 0.640 for the strongest evaluated text-only
baseline. On Falcon-7B, targeted interventions achieve 0.779 AUROC, whereas
shuffling counterfactual assignments reduces performance to 0.533; target
deletion is effective while unrelated deletion is not. However, adjustment for
meaning-preserving paraphrases reduces AUROC to 0.591--0.624.
COVERAGE-TRACE therefore provides a practical, complementary signal of
target-specific source--summary dependence, but not a wording-invariant
semantic coverage probability. These results establish source-side intervention
as a promising approach to omission detection while identifying semantic
invariance as the central remaining challenge.
\end{abstract}

\section{Introduction}
\label{sec:introduction}

A summary can be factually supported yet incomplete because it omits a decision, qualification, numerical detail, or condition needed to preserve the source's meaning. Such omissions are difficult to detect. Unlike hallucinated claims, an omitted fact leaves no output span to verify. Existing factuality metrics such as FactCC~\citep{kryscinski-etal-2020-evaluating}, QAFactEval~\citep{fabbri-etal-2022-qafacteval}, and AlignScore~\citep{zha-etal-2023-alignscore} primarily assess whether generated content is supported by the source. Completeness evaluators instead recover or judge expected information from the response using fact matching, QA, entailment, or LLM judgments~\citep{song-etal-2024-finesure,dejl-etal-2026-comprehensiveness}. Both approaches therefore largely reason from \emph{observed output}, whereas an omission is defined by what is absent.
We instead ask whether the model's support for a summary {depends on a target fact being present in the source}. If that fact is replaced with a plausible alternative, a summary expressing it should exhibit a larger likelihood change than one that omits it. This reframes omission detection as measuring source--summary dependence rather than searching the output for missing content.

We introduce \textbf{COVERAGE-TRACE}, a source-side counterfactual framework that replaces an atomic target fact with controlled alternatives and evaluates the same candidate summary under the original and counterfactual sources using teacher-forced token log-likelihoods. The likelihood change measures dependence on the target fact, while aggregation across multiple valid counterfactuals reduces sensitivity to any single replacement. Because likelihood changes may also reflect lexical or contextual perturbations, we evaluate not only discrimination but also construct validity. Shuffled counterfactuals, unrelated deletions, and meaning-preserving rewrites test target specificity and surface-form sensitivity; text-only evaluators and activation patching provide complementary behavioral and mechanistic comparisons.

On a frozen QMSum-derived benchmark~\citep{zhong-etal-2021-qmsum} of 80 matched covered--omitted pairs, multi-counterfactual COVERAGE-TRACE achieves 0.764--0.825 AUROC across four open-weight models. On Falcon-7B, targeted interventions achieve 0.779 AUROC versus 0.533 when shuffled, while the strongest evaluated text-only method reaches 0.640 compared with 0.825 for multi-counterfactual likelihood tracing. Meaning-preserving rewrites, however, substantially reduce discrimination, showing that the signal remains partly coupled to source wording.

Our contributions are:
\begin{itemize}
    \item We formulate omission evaluation as source--summary dependence on a designated atomic fact rather than recovery of that fact from the response.
    \item We introduce \textbf{COVERAGE-TRACE}, a likelihood-based counterfactual evaluator that aggregates multiple controlled source interventions.
    \item We demonstrate consistent omission discrimination across four open-weight models and show through intervention controls that the signal is target-specific but not invariant to source wording.
\end{itemize}


\section{Related Work}
\label{sec:related_work}

\paragraph{Factuality and coverage evaluation.}
Factuality metrics such as FactCC~\citep{kryscinski-etal-2020-evaluating},
QAFactEval~\citep{fabbri-etal-2022-qafacteval},
AlignScore~\citep{zha-etal-2023-alignscore},
FActScore~\citep{min-etal-2023-factscore}, and
SelfCheckGPT~\citep{manakul-etal-2023-selfcheckgpt} primarily evaluate
information present in a generation. Completeness methods instead assess
whether expected information is preserved, using fine-grained judgments,
fact matching, NLI, QA, or LLM evaluation
~\citep{song-etal-2024-finesure,samarinas-etal-2025-beyond,elaraby2026arc}.
Recent black-box omission and hallucination signals include importance-aware
recall, consistency under uncertain expressions, self-consistency polling,
geometric uncertainty, and embedding-density checking
~\citep{manakul-etal-2023-selfcheckgpt,
friel2023chainpoll,phillips2025geometric,oukelmoun-etal-2025-detecting}.
These methods recover, align, or judge information from the response;
COVERAGE-TRACE instead asks whether a designated \emph{source fact}
measurably influences the response.

\paragraph{Evaluator validity and robustness.}
LLM-based evaluators such as G-Eval~\citep{liu-etal-2023-g} have motivated
work on evaluator reliability and bias~\citep{zheng2023judging}, including
robustness to meaning-preserving reformulations~\citep{bhat-varma-2026-prompts}.
Our validity analysis applies a related principle to the source: we perturb
source evidence while holding the candidate and scoring procedure fixed,
using shuffled, deletion, and paraphrase controls to distinguish
target-specific dependence from generic perturbation sensitivity.

\paragraph{Counterfactual interventions and causal tracing.}
Causal tracing uses internal interventions to identify computations
supporting factual predictions~\citep{meng2022locating}. COVERAGE-TRACE
instead intervenes on the \emph{source text} and measures the resulting
change in support for a fixed candidate, using intervention as an
evaluation signal for coverage. Activation patching is used only as
complementary mechanistic evidence; the omission score itself depends
solely on behavioral responses to source counterfactuals.\\

\section{Methodology}
\label{sec:method}

\subsection{Problem Formulation and Overview}
\label{sec:method_overview}

\begin{figure*}[t]
\centering
\includegraphics[width=1.0\textwidth]{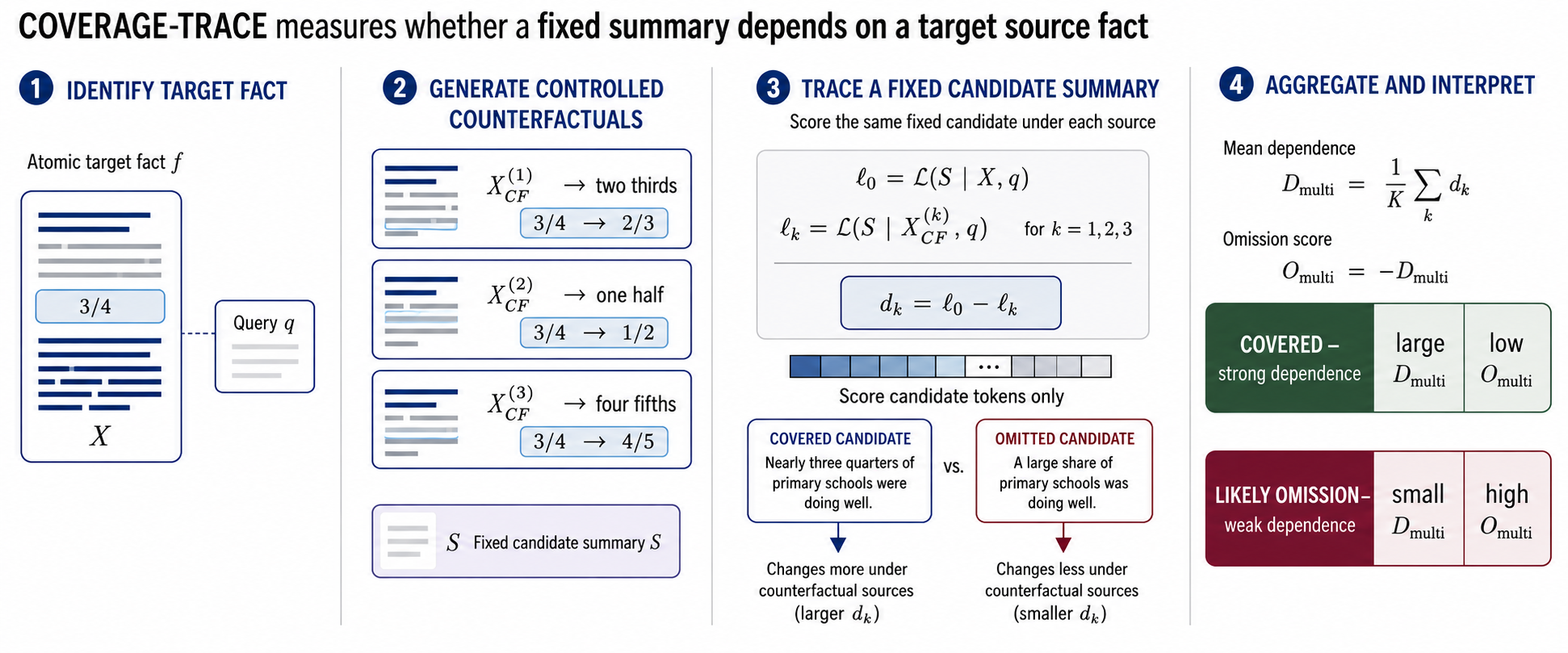}
\caption{COVERAGE-TRACE holds the candidate fixed and measures its likelihood
under the original source and targeted counterfactual sources. Multi-CF
averages intervention effects; larger omission-oriented scores indicate weaker
dependence on the target fact.}
\label{fig:methodology}
\end{figure*}

Let $X$ denote a source document, $q$ a query, $S=(s_1,\ldots,s_T)$ a
candidate summary, and $f$ a designated atomic source fact realized by a span
in $X$. Our goal is to determine whether $S$ covers $f$; selecting which facts
to evaluate is outside the scope of the method.
COVERAGE-TRACE measures coverage from the source side by replacing $f$ with a
valid alternative $f^{(k)}$ to obtain a counterfactual source
$X_{\mathrm{CF}}^{(k)}$, while holding the remaining context and candidate
summary fixed. If $S$ expresses $f$, its model support should change more
strongly under this intervention than if $S$ omits $f$.
Figure~\ref{fig:methodology} illustrates the framework. Our primary estimator,
\textbf{CT-W}, measures this change using teacher-forced token log-likelihoods,
while \textbf{Multi-CF} averages the effect over several valid
counterfactuals.

\subsection{Source Counterfactual Intervention}
\label{sec:counterfactual}

For a target fact $f$, we define
\begin{equation}
X_{\mathrm{CF}}^{(k)}=\mathcal{I}\!\left(X,f;f^{(k)}\right),
\label{eq:cf_source}
\end{equation}
where only the designated realization of $f$ is replaced. A valid
counterfactual must (i) change the target proposition rather than paraphrase
it, (ii) preserve its semantic type and local linguistic form, and (iii)
remain grammatical and contextually plausible. For example,
\begin{quote}
\small
\textit{Nearly three quarters of primary schools performed well.}
$\longrightarrow$
\textit{Nearly two thirds of primary schools performed well.}
\end{quote}
changes the target quantity while preserving the surrounding proposition.
These localized edits are intended to isolate sensitivity to $f$; Section~
\ref{sec:experimental_setup} tests this interpretation with shuffled,
deletion, and paraphrase controls.

\subsection{Likelihood-Based Coverage Tracing}
\label{sec:ctw}

Our primary estimator, \textbf{CT-W}, requires teacher-forced token
log-probabilities but no access to model parameters or internal activations.
For a fixed candidate summary $S=(s_1,\ldots,s_T)$, we define its
length-normalized log-likelihood as
\begin{equation}
    \ell_{\theta}(S\mid X)
    =
    \frac{1}{T}
    \sum_{t=1}^{T}
    \log p_{\theta}
    \left(s_t \mid X,s_{<t}\right).
    \label{eq:summary_likelihood}
\end{equation}
Only candidate-summary tokens contribute to this quantity. The summary is
held fixed: the identical sequence $S$ is scored under the original source
$X$ and counterfactual source $X_{\mathrm{CF}}$ (the Single-CF case of
$X_{\mathrm{CF}}^{(k)}$). Section~\ref{sec:reproducibility} specifies the scoring
prompt and the role of $q$ in the benchmark.

We measure dependence on the target fact by
\begin{equation}
    \Delta_{\theta}(S;f)
    =
    \ell_{\theta}(S\mid X)
    -
    \ell_{\theta}(S\mid X_{\mathrm{CF}}).
    \label{eq:delta}
\end{equation}
If $S$ expresses $f$, replacing $f$ should reduce its likelihood, yielding a
larger positive $\Delta_{\theta}$; a summary that omits $f$ should be less
affected.

Because our detection task treats omission as the positive class, we reverse
the score orientation:
\begin{equation}
    \operatorname{CT\text{-}W}_{\theta}(S;f)
    =
    -\Delta_{\theta}(S;f).
    \label{eq:ctw_score}
\end{equation}
Thus, larger CT-W values indicate greater evidence of omission.
For controlled evaluation with matched covered and omitted summaries, we
additionally define
\begin{equation}
    \Delta_{\mathrm{gap}}
    =
    \Delta_{\theta}(S_{\mathrm{full}};f)
    -
    \Delta_{\theta}(S_{\mathrm{omit}};f).
    \label{eq:delta_gap}
\end{equation}
The hypothesis predicts $\Delta_{\mathrm{gap}}>0$, meaning that intervention
on $f$ affects the covered summary more strongly than its matched omitted
counterpart. This paired quantity is used only for analysis; individual
candidates are scored directly with CT-W.

\subsection{Multi-Counterfactual Tracing}
\label{sec:multi_cf}

A Single-CF score may depend on the particular replacement chosen for $f$.
Multi-CF reduces this dependence by aggregating responses across $K_f$ valid
counterfactuals. For the $k$th counterfactual source
$X_{\mathrm{CF}}^{(k)}$, we compute
\begin{equation}
\Delta_\theta^{(k)}(S;f)
=
\ell_\theta(S\mid X)
-
\ell_\theta(S\mid X_{\mathrm{CF}}^{(k)}),
\end{equation}
and average the responses:
\begin{equation}
\overline{\Delta}_{\theta}(S;f)
=
\frac{1}{K_f}
\sum_{k=1}^{K_f}
\Delta_\theta^{(k)}(S;f).
\label{eq:multi_delta}
\end{equation}
Following the Single-CF orientation, the Multi-CF omission score is
\begin{equation}
\operatorname{CT\text{-}W}_{\mathrm{Multi},\theta}(S;f)
=
-\overline{\Delta}_{\theta}(S;f).
\label{eq:multi_score}
\end{equation}
Thus, Multi-CF estimates the same source--summary dependence as Single-CF
while reducing reliance on any single counterfactual.

\section{Experimental Setup}
\label{sec:experimental_setup}

We derive 80 matched covered--omitted pairs
(160 candidates) from 68 distinct QMSum source
documents \citep{zhong-etal-2021-qmsum}.
Each item specifies a source, query, target fact,
original reference, controlled omission preserving
other content, and one to three localized
counterfactuals.
The targets include 24 decision rationales,
24 qualifications, 20 quantities, and 12 temporal,
causal, or comparative facts.
Candidate omissions and source-side counterfactuals
are constructed using explicit rule-based
transformations rather than LLM-generated labels.
Records pass source-support, lineage, and targeting
checks. For likelihood evaluation, each source is restricted
to a 3,500-character scoring window. An automated
integrity check confirms that the designated target
cue remains visible within this window for all
80 matched pairs.
All scoring models remain frozen during evaluation;
COVERAGE-TRACE requires no additional training
or fine-tuning.
Neither exact matching (0.512 AUROC) nor
target-token recall (0.609 AUROC) trivially
solves the task.
\paragraph{Models and scoring.}
We evaluate Falcon-7B-Instruct, a NousResearch-distributed
Llama-3-8B-Instruct checkpoint, Qwen2.5-7B-Instruct,
and Qwen2.5-14B-Instruct.
The fixed prompt presents the source and requests
a faithful summary; the identical candidate is
scored under original and counterfactual sources.
AUROC uses omission as the positive class.

Confidence intervals and method differences are
estimated using 10,000 paired bootstrap resamples
with replacement, preserving the multiplicity
of sampled observations.
Baselines include DeBERTa atomic NLI,
a target-level importance-aware recall adaptation,
a DistilBERT-NLI alignment proxy, and locally
implemented consistency- and uncertainty-based
evaluators.

Three controls probe construct validity:
shuffling counterfactual assignments tests
fact--intervention alignment; target versus
unrelated deletion tests target-span specificity;
and local meaning-preserving rewrites test
surface-form sensitivity. For the latter,
we subtract the average likelihood change induced
by meaning-preserving rewrites from the likelihood
change induced by target-changing interventions.

\section{Results}
\paragraph{Multi-CF consistently improves discrimination.}
Table~\ref{tab:cross_model} shows gains of 0.046--0.051 AUROC across all four
models. On Falcon-7B, the paired improvement is 0.046 (95\% CI [0.022, 0.073]);
the hypothesized dependence gap is positive for 68/80 pairs under Single-CF
and 72/80 under Multi-CF. Multi-CF reaches 0.825 AUROC, versus 0.640 for the
strongest evaluated baseline, DeBERTa atomic NLI; other baselines score
0.501--0.622.
Representative qualitative examples illustrating
source-side interventions and their effects on
omission discrimination are provided in
Appendices~\ref{app:qualitative} and~\ref{app:qualitative_cases}.
\begin{table}[h]
\centering\small
\caption{COVERAGE-TRACE AUROC (95\% paired-bootstrap CI).}
\vspace{5 pt}
\label{tab:cross_model}
\begin{tabular}{lccc}\toprule
Model & Single-CF & Multi-CF & Gain\\\midrule
Falcon-7B & .779 [.720,.840] & \textbf{.825 [.772,.879]} & +.046\\
Llama-3-8B & .756 [.699,.813] & \textbf{.802 [.747,.855]} & +.046\\
Qwen2.5-7B & .713 [.650,.777] & \textbf{.764 [.706,.823]} & +.051\\
Qwen2.5-14B & .764 [.708,.822] & \textbf{.814 [.762,.868]} & +.049\\\bottomrule
\end{tabular}
\end{table}
\paragraph{Statistical reliability of aggregation.}
The improvement from Single-CF to Multi-CF
is consistently positive across all four
scoring models, with paired AUROC gains of
0.046, 0.046, 0.051, and 0.049 for Falcon-7B,
Llama-3-8B, Qwen2.5-7B, and Qwen2.5-14B,
respectively.
The corresponding paired-bootstrap 95\%
confidence intervals are [0.022, 0.073],
[0.016, 0.079], [0.019, 0.088], and
[0.016, 0.085], respectively.
All four intervals exclude zero, providing
consistent statistical evidence for the
benefit of counterfactual aggregation.

\paragraph{The signal is target-specific but surface-sensitive.}
Table~\ref{tab:validity_controls} summarizes the
intervention controls across all four scoring models.
Shuffling counterfactual assignments substantially
reduces omission discrimination. Targeted Single-CF
achieves AUROC values of 0.713--0.779, whereas
shuffled interventions yield only 0.533--0.553.
On Falcon-7B, AUROC decreases from 0.779 to 0.533
(paired difference 0.245; 95\% CI [0.184, 0.303]).
The same pattern holds for Llama-3-8B,
Qwen2.5-7B, and Qwen2.5-14B, with paired
targeted-minus-shuffled improvements of
0.204, 0.170, and 0.231, respectively.
Their paired-bootstrap 95\% confidence intervals are
[0.143, 0.254], [0.105, 0.229], and
[0.169, 0.285], respectively.
These results demonstrate that the discriminative
signal depends on aligning counterfactual
interventions with the designated source fact,
rather than on arbitrary source perturbations.


\begin{table}[h]
\centering
\small
\caption{Intervention-control AUROC across four models. Targeted interventions outperform shuffled controls, while unrelated deletions remain near chance. Paraphrase-adjusted AUROC uses 76 valid pairs; four lacked usable paraphrases.}
\label{tab:validity_controls}
\begin{tabular}{lcccc}
\toprule
Score & Falcon & Llama & Qwen-7B & Qwen-14B \\
\midrule
Targeted CF         & .779 & .756 & .713 & .764 \\
Shuffled CF         & .533 & .553 & .543 & .533 \\
Target deletion     & .782 & .787 & .731 & .764 \\
Random deletion     & .493 & .503 & .502 & .527 \\
Paraphrase-adjusted ($n=76$) & .610 & .594 & .591 & .624 \\
\bottomrule
\end{tabular}%
\end{table}

Target deletion performs comparably to targeted
counterfactual replacement across all four models,
with AUROC values of 0.731--0.787, whereas
random deletion remains near chance
(0.493--0.527). These controls provide additional evidence that
the signal is associated with perturbations of
the designated fact-bearing source span.
However, they do not establish a clear advantage
of semantic replacement over simply removing
the target information.
Meaning-preserving rewrites reveal an important
construct-validity limitation: the score is not
invariant to surface form.
Paraphrase-adjusted AUROC decreases to 0.610
on Falcon-7B and 0.591--0.624 on the other
models (76 valid pairs; four lacked usable
paraphrases). Raw Single-CF AUROC is reported on all 80 pairs, so the comparison is not matched. Across models, responses to target-changing
and meaning-preserving interventions are
moderately correlated
(Pearson $r=.355$; Spearman $\rho=.416$).
Thus, COVERAGE-TRACE captures target-specific
source--summary dependence, but the measured
signal remains partly sensitive to lexical
variation rather than exclusively reflecting
semantic coverage.


\subsection{Aggregation, Baselines, and Localization}
\label{sec:extended_analysis}

\paragraph{Effect of counterfactual aggregation.}
We first investigate how the number of counterfactual
interventions influences omission discrimination and
sensitivity to replacement selection. Let $K$ denote the number of counterfactual alternatives
used to evaluate each target fact. For $K=1$ and $K=2$,
we sample available alternatives without replacement
and repeat the evaluation across 20 random seeds.
For $K=3$, all available alternatives are included,
eliminating randomness associated with counterfactual
selection.
As shown in Table~\ref{tab:cf_budget}, mean Falcon-7B
AUROC increases from 0.809 with one counterfactual
to 0.820 with two and 0.825 with three.
The largest incremental improvement occurs when
increasing $K$ from one to two (+0.011 AUROC),
whereas introducing the third counterfactual provides
a smaller additional gain (+0.005 AUROC).
More importantly, aggregation reduces sensitivity
to the particular counterfactual selected.
The standard deviation of AUROC across random
selections decreases from 0.0125 for $K=1$
to 0.0065 for $K=2$.
For $K=3$, selection variance becomes zero because
all available alternatives are used.
This does not imply zero prediction uncertainty;
rather, the resulting score is deterministic
with respect to counterfactual selection.
These findings indicate that aggregating multiple
interventions improves both discrimination and
stability, with the second counterfactual providing
the largest incremental benefit in this experiment.

\begin{table}[h]
\centering
\small
\caption{
Effect of the number of counterfactual interventions
on Falcon-7B omission discrimination.
Selection SD measures variability across 20 random
counterfactual selections, rather than uncertainty
in individual predictions.
}
\vspace{- 7 pt}
\label{tab:cf_budget}
\begin{tabular}{cccc}
\toprule
$K$ & Mean AUROC & Selection SD & Seed range \\
\midrule
1 & .809 & .0125 & [.794,.839] \\
2 & .820 & .0065 & [.813,.833] \\
3 & \textbf{.825} & .0000 & [.825,.825] \\
\bottomrule
\end{tabular}
\end{table}

\paragraph{Original-source likelihood baseline.}
To isolate the contribution of source-side
interventions, we evaluate a likelihood-only
baseline that scores each candidate summary
under the original source without applying
any counterfactual modification.
The omission-oriented score is
$s_{\mathrm{LL}}(S,X)=-\ell_{\theta}(S\mid X)$,
where $\ell_{\theta}$ denotes the mean
teacher-forced token log-likelihood.
As shown in Table~\ref{tab:likelihood_baseline},
the likelihood-only baseline achieves
0.581--0.603 AUROC across the four models,
substantially below Multi-CF (0.764--0.825).

\begin{table}[h]
\centering
\small
\caption{
Comparison of original-source likelihood
and Multi-CF omission discrimination.
$\Delta$ denotes the paired AUROC difference
(Multi-CF minus likelihood-only).
Confidence intervals are obtained using
10,000 paired bootstrap resamples.
}
\vspace{- 5pt}
\label{tab:likelihood_baseline}
\begin{tabular}{lcccc}
\toprule
Model & LL & Multi-CF & $\Delta$ & 95\% CI \\
\midrule
Falcon-7B
& .581 & .825 & +.244 & [.181,.300] \\
Llama-3-8B
& .601 & .802 & +.201 & [.141,.251] \\
Qwen2.5-7B
& .591 & .764 & +.173 & [.111,.226] \\
Qwen2.5-14B
& .603 & .814 & +.211 & [.148,.264] \\
\bottomrule
\end{tabular}
\end{table}

The paired improvements range from
0.173 to 0.244 AUROC, with
95\% bootstrap confidence intervals
excluding zero in every case.
These findings indicate that source-side
counterfactual interventions provide
discriminative information beyond
unmodified conditional likelihood
under the evaluated protocol.

\paragraph{Comparison with baseline evaluators.}
We compare COVERAGE-TRACE against eight baseline
evaluators representing entailment-based, recall-based,
consistency-based, and uncertainty-based approaches.
All scores are oriented such that higher values
indicate stronger evidence of omission.

The strongest evaluated baseline is DeBERTa atomic
NLI (AUROC 0.640). IAR-target (0.622) is a local
target-level implementation inspired by
importance-aware recall. It combines a binary
COVERED/NOTCOVERED judgment from Qwen2.5-14B
with the entailment probability from
DeBERTa-v3-base-mnli-fever-anli, using the
candidate summary as premise and the designated
fact as hypothesis. The omission score is
one minus the mean of these two coverage signals.
Unlike the original importance-aware recall
framework, this implementation does not use
external retrieval or a multi-fact importance
aggregation.
The DistilBERT-NLI alignment proxy (0.594) uses
DistilBERT-MNLI entailment probabilities
between the candidate summary and target fact,
with omission score $1-P_{\mathrm{entail}}$.
This is not an evaluation of the official
AlignScore checkpoint, which was unavailable
in the experimental environment.

\begin{figure*}[h]
\centering
\includegraphics[
width=0.96\columnwidth
]{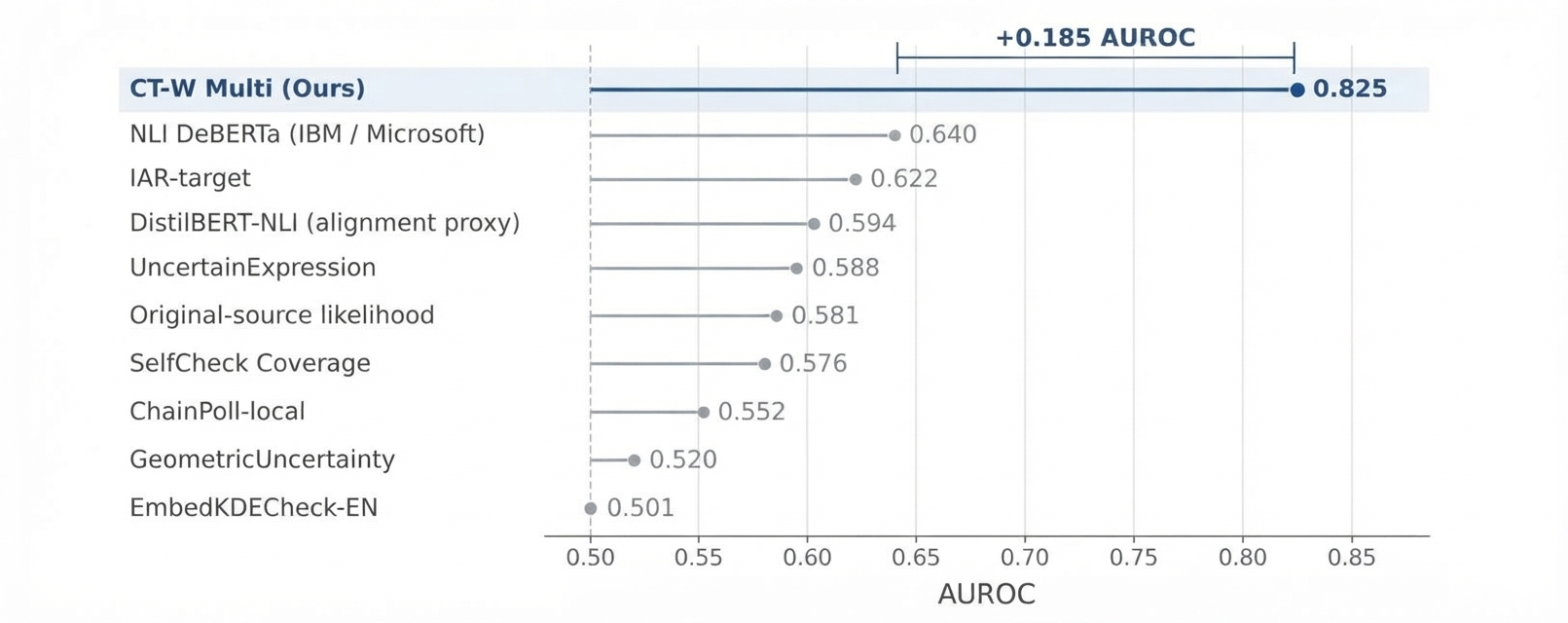}
\vspace{- 8 pt}
\caption{
Omission-discrimination AUROC on Falcon-7B.
COVERAGE-TRACE with multiple counterfactuals
achieves 0.825 AUROC, exceeding the strongest
evaluated alternative, DeBERTa atomic NLI,
by 0.185 AUROC.
Original-source likelihood achieves 0.581 AUROC,
indicating a 0.244 AUROC advantage for Multi-CF
over scoring without source interventions.
Several comparison methods are local adaptations
or proxies rather than official implementations.
The dashed line marks chance performance
(AUROC = 0.5).
}
\label{fig:baseline_comparison}
\end{figure*}

ChainPoll-local (0.552) is an adapted
open-weight consistency baseline. It samples
eight omission judgments from Qwen2.5-14B
at temperature 0.5, using the source document
and candidate summary as input. Its score is
the proportion of responses indicating an
omission, rather than the original GPT-based
ChainPoll procedure.
The remaining evaluated methods include
UncertainExpression (0.588), SelfCheck Coverage
(0.576), GeometricUncertainty (0.520), and
EmbedKDECheck-EN (0.501). As shown in Figure~\ref{fig:baseline_comparison},
Multi-CF achieves 0.825 AUROC, exceeding the
strongest evaluated baseline by 0.185. The advantage over the evaluated DeBERTa
atomic-NLI baseline is not confined to Falcon-7B.
Multi-CF achieves additional paired AUROC
improvements of 0.163, 0.124, and 0.174
on Llama-3-8B, Qwen2.5-7B, and
Qwen2.5-14B, respectively.
The corresponding paired-bootstrap 95\%
confidence intervals are [0.096, 0.227],
[0.056, 0.190], and [0.114, 0.235],
all excluding zero.
The advantage over the evaluated
atomic-NLI baseline persists across the
four scoring models, although the methods
differ in their model-access requirements.
These results demonstrate the discriminative
value of source-side interventions under the
controlled evaluation protocol. However, several
baselines are local adaptations or proxies,
and COVERAGE-TRACE requires teacher-forced
token probabilities unavailable to some
comparison methods. The results therefore
should not be interpreted as establishing
universal superiority over the original
published systems.

\paragraph{Activation-patching localization analysis.}
To investigate whether the source-side omission
signal is associated with representations at the
designated fact-bearing positions, we conduct
a complementary residual-stream activation-patching
analysis on Falcon-7B.
For each example, we identify the contiguous
source-token span modified by the counterfactual
intervention. We restrict the analysis to 57
covered--omitted pairs for which the original
and counterfactual sources have exactly aligned
tokenizations.
We first perform a forward pass using the
original source and record the residual-stream
activations at the intervention positions.
We then evaluate the counterfactual source,
replacing the corresponding residual-stream
vectors with their original-source counterparts
at a selected transformer layer.
Patching is performed on the full hidden-state
vector at the output of each FalconDecoderLayer,
after the attention and MLP computations.
The candidate summary remains unchanged and
is evaluated using teacher-forced token
log-likelihoods.
As a localization control, we repeat the
procedure at an unrelated source span of
identical token length. The control span
contains unchanged tokens and is selected
deterministically outside the intervention
region.
For a candidate summary $S$, we define
the likelihood restoration at layer $\ell$ as
\begin{equation}
R_{\ell}(S)
=
\mathcal{L}_{\mathrm{patched},\ell}(S)
-
\mathcal{L}(S\mid X_{\mathrm{CF}}),
\end{equation}

where $\mathcal{L}$ denotes the mean
teacher-forced token log-likelihood.

The patching discrimination gap is

\begin{equation}
G_{\ell}
=
R_{\ell}(S_{\mathrm{full}})
-
R_{\ell}(S_{\mathrm{omit}}).
\end{equation}

We define localization gain as the difference
between the discrimination gaps obtained
from target-span and control-span patching:

\begin{equation}
\mathrm{LG}_{\ell}
=
G_{\ell}^{\mathrm{target}}
-
G_{\ell}^{\mathrm{control}}.
\end{equation}

\begin{figure*}[b]
\centering
\includegraphics[
    width=0.7\textwidth
]{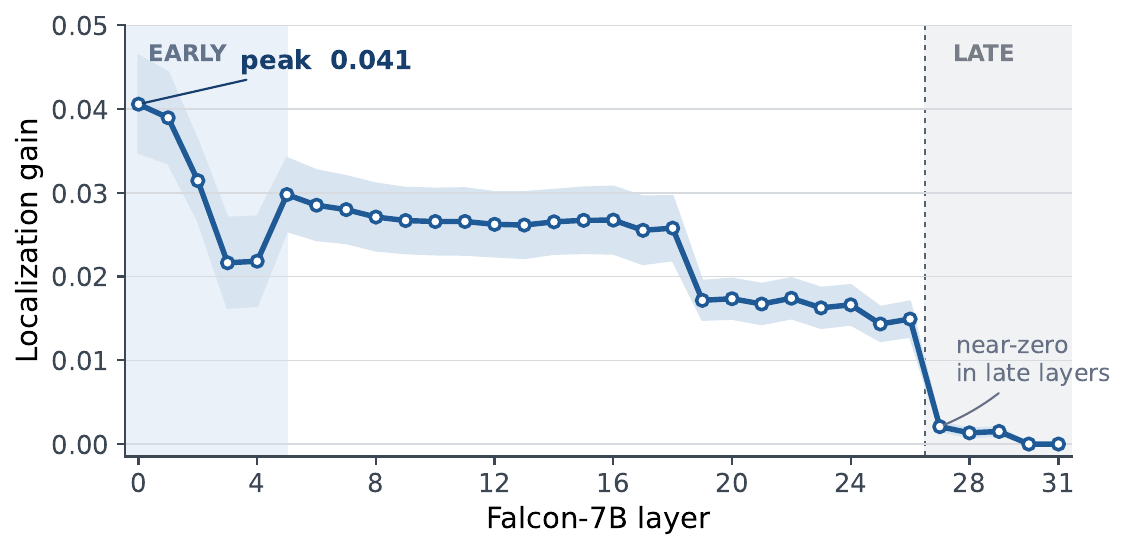}
\vspace{- 5 pt}

\caption{
\textbf{Layer-wise localization gain on Falcon-7B.}
We perform clean-to-counterfactual residual-stream
activation patching across all 32 transformer layers
on 57 token-aligned covered--omitted pairs.
Localization gain measures the difference between
covered-minus-omitted likelihood restoration
under target-span and same-length control-span
patching. Shading indicates $\pm 1$ SEM.
The gain peaks at layer 0 (0.041), remains positive
through intermediate layers, and approaches zero
in the final layers.
}

\label{fig:depth_localization}
\end{figure*}

A positive localization gain indicates
that restoring clean activations at the
fact-bearing source positions preferentially
restores the likelihood of the covered summary,
relative to the omitted summary, more strongly
than restoring activations at unrelated positions.
Figure~\ref{fig:depth_localization} reports
the mean localization gain across the 57
aligned pairs for all 32 Falcon-7B layers.
The gain is largest at layer 0
(approximately 0.041), remains positive
through intermediate layers, and decreases
toward zero in the final transformer blocks.
These results indicate that the evaluated
patching intervention has a stronger
target-specific effect in earlier layers.
However, the layer-dependent patching response
does not establish where factual information
is exclusively represented or identify a
unique causal circuit.
The analysis provides complementary evidence
of sensitivity to fact-bearing source
representations; the COVERAGE-TRACE omission
score itself does not require access to
internal activations.

\paragraph{Error characterization.}
Our method generalizes beyond quantitative facts. Single-CF achieves AUROC
0.888 on quantitative details ($n=20$), 0.766 on qualifications ($n=24$),
and 0.719 on decision rationales ($n=24$). The remaining 12temporal, causal, and comparative cases are not analyzed separately because they form a small, heterogeneous group. Because these subsets are small,
we report them descriptively rather than as statistically significant
comparisons. The method is robust to the choice of valid counterfactual. Across
retained counterfactual sets, 90\% preserve the sign of the paired dependence
gap. For example, replacing ``nearly three quarters'' with three
alternatives yields a Multi-CF gap of 0.081 and the correct ordering under all
four CT-W models. In the object/material example, replacing ``titanium'' with
\{plastic, aluminum, steel\} yields a Multi-CF gap of 0.150 and the correct
ordering under all four CT-W models. DeBERTa NLI and IAR-target also rank both
examples correctly; these cases illustrate target-specific source dependence
without claiming a semantic-baseline reversal.

\vspace{- 6 pt}
\subsection{Construction and Reproducibility}
\label{sec:reproducibility}

\paragraph{Benchmark construction.}
Each benchmark record links an explicit source span
to an atomic target fact, a covered reference,
a minimally edited omitted candidate, and
localized counterfactuals.
We reject replacements unsupported by the edited
source, changes affecting propositions other than
the target, exact-string leakage, generic omission
templates, and mismatched source--fact associations.
Retained records pass source-support, lineage,
and targeting checks.
This controlled design isolates the effect of
targeted source interventions while separating
omission detection from the distinct task of
determining which source facts warrant inclusion.

\vspace{- 6 pt}
\paragraph{Likelihood scoring.}
The prompt is fixed across models: the source
is prefixed with ``Source document:'' and followed
by ``Write a faithful summary of the source.''
The identical candidate summary is appended and
scored using teacher forcing under both original
and counterfactual sources. The loss mask excludes
the prompt and source, so likelihood is computed
only over candidate-summary tokens.
Although QMSum provides a query for each item,
we do not separately include it in the scoring
prompt. This design isolates source-conditioned
coverage rather than query-conditioned
summarization.
Single-CF requires two likelihood evaluations
per candidate. Multi-CF requires $1+K_f$
evaluations, reusing the original-source score.

\vspace{- 6 pt}
\paragraph{Validity controls.}
All validity controls preserve the same
likelihood-scoring pipeline.
Counterfactual shuffling changes only the
example-to-counterfactual assignment, testing
whether the observed signal depends on the
correct fact--intervention association.
Deletion controls compare removal of the
designated target fact against removal of
an unrelated local source span, testing
target-span specificity.
For paraphrase-adjusted AUROC, we use the 76 pairs with
usable meaning-preserving rewrites; four of the original
80 pairs lack a usable paraphrase and are excluded from
this control only.
Finally, paraphrase adjustment subtracts
the average likelihood effect of local
meaning-preserving rewrites from the effect
of fact-changing interventions before
reversing the score for omission detection. Its performance decline indicates
that raw CT-W is not invariant to
meaning-preserving wording changes, rather
than invalidating the observed discrimination
under the original evaluation protocol.

\vspace{- 7 pt}
\section{Limitations}
\label{app:limitations}

COVERAGE-TRACE remains sensitive to source wording, with paraphrasing reducing AUROC to 0.591--0.624, limiting its interpretation as a purely semantic coverage measure. Evaluation is restricted to 80 controlled QMSum pairs, including 12 with construction defects, although the main aggregation benefit persists after their exclusion. The method also requires access to model likelihoods, while activation-patching experiments were limited to Falcon-7B. 

\vspace{- 7 pt}
\section{Conclusion}
\label{sec:conclusion}

We introduced COVERAGE-TRACE, a source-side
counterfactual framework for detecting omissions
in LLM-generated summaries. By measuring how
targeted changes to source facts affect the
likelihood of a fixed candidate summary, the
method identifies source-summary dependencies
without requiring additional model training.
Experiments across four open-weight language
models demonstrate consistent omission
discrimination, with Multi-CF achieving
0.764--0.825 AUROC and outperforming
the evaluated baselines. Intervention controls
further support the importance of targeting
the correct source facts.
While the approach remains sensitive to
surface wording and is evaluated in a
controlled setting, it provides a promising
direction for detecting fact-level omissions.


\clearpage
\bibliographystyle{plainnat}
\bibliography{references}
\clearpage

\section*{APPENDIX}




\section{Benchmark Integrity Sensitivity Analysis}
\label{app:integrity}

We conduct an additional sensitivity analysis
to assess whether identifiable mechanical
benchmark-construction defects account for
the observed performance improvements.
An automated integrity audit flagged
12 of the original 80 matched pairs
for concrete construction defects,
including word-internal substitutions
and duplicated target propositions
left unchanged by the intervention.
We exclude these 12 pairs and recompute
the evaluation using the previously
stored model scores.
The resulting subset contains 68 matched
pairs from 57 distinct source documents.
No model parameters, scoring procedures,
or counterfactual definitions are modified.
Table~\ref{tab:integrity_results} reports
the resulting Multi-CF performance and
paired improvements over Single-CF.
\begin{table}[h]
\centering
\small
\caption{
Integrity-screened robustness analysis
on 68 matched pairs from 57 documents.
$\Delta$ denotes Multi-CF minus Single-CF
AUROC. Confidence intervals use
paired bootstrap resampling.
}
\label{tab:integrity_results}

\begin{tabular}{lccc}
\toprule
Model &
Multi-CF &
$\Delta$ &
95\% CI \\
\midrule
Falcon-7B
& .821 & +.044 & [.018,.074] \\
Llama-3-8B
& .806 & +.046 & [.015,.084] \\
Qwen2.5-7B
& .749 & +.050 & [.014,.095] \\
Qwen2.5-14B
& .805 & +.052 & [.016,.091] \\
\bottomrule
\end{tabular}
\end{table}
The Multi-CF advantage remains positive
for every scoring model after excluding
the identified defective constructions.
However, this experiment constitutes
a within-benchmark sensitivity analysis,
not an independent evaluation.
The integrity audit additionally identified
16 cases requiring further human review;
therefore, the retained subset should
not be interpreted as fully human-validated.

\vspace{- 6 pt}

\section{Qualitative Examples}
\label{app:qualitative}
Table~\ref{tab:qualitative} illustrates the intervention signal on two
benchmark examples: a quantitative-detail case and an object/material case.
In both, Multi-CF CT-W and the evaluated semantic coverage baselines correctly
separate the covered and omitted candidates.
\begin{table*}[ht]
    \centering
    \small
    \setlength{\tabcolsep}{3pt}
    \caption{
        \textbf{Qualitative examples from the controlled benchmark.}
        The original QMSum reference is used as $S_{\mathrm{full}}$.
        Only locally relevant excerpts are shown. An evaluator is correct
        when it assigns greater omission evidence to $S_{\mathrm{omit}}$
        than to $S_{\mathrm{full}}$.
    }
    \label{tab:qualitative}
    \begin{tabular}{
        p{0.16\textwidth}
        p{0.25\textwidth}
        p{0.30\textwidth}
        p{0.22\textwidth}}
        \toprule
        \textbf{Type}
        & \textbf{Original QMSum item}
        & \textbf{Controlled construction}
        & \textbf{Evaluator behavior} \\
        \midrule

        \textbf{Quantitative detail}
        &
        \textbf{Query:}
        To what extent had improvement differed between the primary and
        secondary sectors?

        \textbf{Source:}
        ``We've got 7 in 10---nearly \emph{three quarters}---of primary
        schools doing well and about half of secondary schools.''

        \textbf{Reference:}
        ``Nearly \emph{three quarters} of primary schools were doing well
        \ldots''
        &
        \textbf{Target fact:}
        Nearly three quarters of primary schools were doing well.

        \textbf{$S_{\mathrm{full}}$:}
        ``Nearly \emph{three quarters} of primary schools were doing well
        \ldots''

        \textbf{$S_{\mathrm{omit}}$:}
        ``A \emph{large share} of primary schools were doing well \ldots''

        \textbf{Counterfactuals:}
        three quarters $\rightarrow$
        \{two thirds, one half, one third\}
        &
        \textbf{Multi-CF CT-W:}
        $\Delta_{\mathrm{gap}}=\mathbf{0.081}$; correct (4/4 models).
        DeBERTa atomic NLI: correct. IAR-target: correct.
        \\
        \midrule

        \textbf{Object / material}
        &
        \textbf{Query:}
        What did the marketing learn about the trend?

        \textbf{Source:}
        ``And not those futuristic uh remote control with angles and uh
        and \emph{titanium} like.''

        \textbf{Reference:}
        ``It might be surprising that people liked products that looked
        natural and round rather than some futuristic remote control with
        angles and \emph{titanium}.''
        &
        \textbf{Target fact:}
        The futuristic remote control was described as having angles and
        titanium.

        \textbf{$S_{\mathrm{full}}$:}
        ``\ldots rather than some futuristic remote control with angles
        and \emph{titanium}.''

        \textbf{$S_{\mathrm{omit}}$:}
        ``\ldots rather than some futuristic remote control with angles
        and \emph{lighter alloy}.''

        \textbf{Counterfactuals:}
        titanium $\rightarrow$
        \{plastic, aluminum, steel\}
        &
        \textbf{Multi-CF CT-W:}
        $\Delta_{\mathrm{gap}}=\mathbf{0.150}$; correct (4/4 models).
        DeBERTa atomic NLI: correct. IAR-target: correct.
        \\
        \bottomrule
    \end{tabular}
\end{table*}
For the quantitative example, replacing ``nearly three quarters'' with
\{two thirds, one half, one third\} yields a Multi-CF gap of 0.081. In the
object/material example, replacing ``titanium'' with
\{plastic, aluminum, steel\} yields a gap of 0.150. All four CT-W models,
DeBERTa atomic NLI, and IAR-target rank both pairs correctly. These examples
show that the source-conditioned likelihood response tracks the designated
fact across distinct omission types.

\section{Qualitative Analysis}
\label{app:qualitative_cases}

We examine representative benchmark cases to
characterize when source-side counterfactual
tracing succeeds and fails. Table~\ref{tab:qualitative_cases}
reports the paired dependence gap
$\Delta_{\mathrm{gap}}$ for selected Falcon-7B
examples. Positive values indicate that
the covered candidate responds more strongly
to the intervention than its omitted counterpart.

\begin{table*}[t]
\centering
\small
\caption{
Representative qualitative cases from Falcon-7B.
Single-CF, Multi-CF, and deletion columns report
paired dependence gaps.
Positive values indicate correct ordering
of the covered and omitted candidates.
Examples are illustrative rather than
statistically representative.
}
\label{tab:qualitative_cases}

\resizebox{\textwidth}{!}{
\begin{tabular}{p{2.0cm}p{3.7cm}rrrp{4.8cm}}
\toprule
Case & Target and candidate contrast
& Single-CF & Multi-CF & Deletion
& Interpretation \\
\midrule

Aggregation benefit
& ``the internet'' vs.\ ``online access''
& $-0.00169$ & $+0.02351$ & $+0.02255$
& Multi-CF reverses an incorrect Single-CF
ordering, although the lexical distinction
is semantically ambiguous. \\

Replacement benefit
& ``six years'' vs.\ ``several years''
& $+0.01906$ & $+0.01906$ & $-0.00580$
& Targeted replacement succeeds where
target deletion fails. \\

Quantitative success
& ``three quarters'' vs.\ ``a large share''
& $+0.07690$ & $+0.08094$ & $+0.06932$
& All three intervention strategies
correctly distinguish the candidates. \\

Temporal challenge
& ``two years'' vs.\ ``several years''
& $-0.00148$ & $-0.00148$ & $-0.00412$
& The interventions fail to recover
the expected ordering. \\

\bottomrule
\end{tabular}
}
\end{table*}

\paragraph{Benefits of counterfactual aggregation.}
In one example, replacing ``the internet''
with alternatives including ``the web'',
``online access'', and ``broadband''
changes the paired dependence gap from
$-0.00169$ under Single-CF to $+0.02351$
under Multi-CF.
This illustrates how averaging multiple
interventions can stabilize an otherwise
incorrect ranking. However, the distinction
between ``the internet'' and ``online access''
is semantically ambiguous, so this example
should not be interpreted as an unambiguous
semantic-omission detection success.
\paragraph{Replacement versus deletion.}
For a temporal fact specifying ``six years'',
the omitted candidate instead uses
``several years''. Replacing the source
duration with ``four years'' produces
a positive dependence gap of $0.01906$,
whereas deleting the target yields
a negative gap of $-0.00580$.
This example illustrates a case in which
targeted replacement succeeds while deletion
does not. It does not contradict the aggregate
finding that deletion performs comparably
to replacement across the benchmark.

\paragraph{Surface-form sensitivity.}
A separate decision-related example reveals
a limitation of the method. The original
target-changing intervention
(``simple'' $\rightarrow$ ``complex'')
produces a positive dependence gap of
$0.04624$. However, replacing ``simple'' with the
meaning-preserving alternative
``straightforward'' changes the gap
to $-0.00584$. Thus, a local paraphrase can reverse
the ordering of the covered and omitted
candidates. Because the paraphrase was
generated automatically, this example
is illustrative rather than a
human-validated semantic-invariance test.

\paragraph{Failure on underspecified durations.}
For another temporal example, the covered
candidate specifies ``two years'', whereas
the omitted candidate uses ``several years''.

Replacing the source duration with
``four years'' yields a negative gap of
$-0.00148$ under both Single-CF and
Multi-CF. Target deletion also fails
($-0.00412$). This case illustrates that source-side
likelihood dependence does not always
distinguish a specific factual value
from a less precise expression.

\end{document}